\documentclass[transaction]{IEEEtran}
\usepackage{amsfonts}
\usepackage{amssymb}
\usepackage{hyperref}
\usepackage{graphicx}
\usepackage{enumerate}
\usepackage{amsmath}
\usepackage{color}
\usepackage{amsthm}
\usepackage{verbatim}
\usepackage{amsmath}
\usepackage{algorithm}
\usepackage{cite}
\usepackage{algpseudocode}
\usepackage{subfig}
\usepackage{graphicx}
\usepackage{float}
\usepackage{autobreak}
\usepackage{booktabs}  
\usepackage{array}   
\usepackage{multirow}
\usepackage{booktabs}
\usepackage{xcolor}  
\usepackage{amsfonts,amssymb}
\providecommand{\keywords}[1]{\textbf{\textit{Index terms---}} #1}
\newcommand{\bp}{\begin{proof} \small }
\newcommand{\ep}{\end{proof} \normalsize}
\newcommand{\epx}{\end{proof} \small}
\newcommand{\bpa}{\begin{proofappx} \footnotesize }
\newcommand{\epa}{\end{proofappx} \small }

\newtheorem*{theorem*}{Theorem}
\newtheorem*{proposition*}{Proposition}
\newtheorem*{corollary*}{Corollary}
\newtheorem*{lemma*}{Lemma}
\newtheorem*{assumption*}{Assumption}
\newtheorem*{definition*}{Definition}
\newtheorem*{claim*}{Claim}

\newcommand{\be}{\begin{equation}}
\newcommand{\ee}{\end{equation}}
\newcommand{\bs}{\begin{subequations}}
\newcommand{\es}{\end{subequations}}
\newcommand{\bq}{\begin{eqnarray}}
\newcommand{\eq}{\end{eqnarray}}
\newcommand{\bqn}{\begin{eqnarray*}}
\newcommand{\eqn}{\end{eqnarray*}}

\newcommand{\ba}{\left[ \begin{array}}
\newcommand{\ea}{\\ \end{array} \right]}
\newcommand{\ben}{\begin{enumerate}}
\newcommand{\een}{\end{enumerate}}
\def\real{{\mathchoice%
{\hbox{\rm\setbox1=\hbox{I}\copy1\kern-.45\wd1 R}}
{\hbox{\rm\setbox1=\hbox{I}\copy1\kern-.45\wd1 R}}
{\hbox{\scriptsize\rm\setbox1=\hbox{I}\copy1\kern-.45\wd1 R}}
{\hbox{\scriptsize\rm\setbox1=\hbox{I}\copy1\kern-.45\wd1 R}}}}

\def\Zint{{\mathchoice{\setbox1=\hbox{\sf Z}\copy1\kern-.75\wd1\box1}
{\setbox1=\hbox{\sf Z}\copy1\kern-.75\wd1\box1}
{\setbox1=\hbox{\scriptsize\sf Z}\copy1\kern-.75\wd1\box1}
{\setbox1=\hbox{\scriptsize\sf Z}\copy1\kern-.75\wd1\box1}}}
\newcommand{\complex}{ \hbox{\rm C\kern-0.45em\rule[.07em]{.02em}{.58em}%
\kern 0.43em}}

\allowdisplaybreaks

\begin{document}
%
% paper title
% can use linebreaks \\ within to get better formatting as desired

%
%
% author names and IEEE memberships
% note positions of commas and nonbreaking spaces ( ~ ) LaTeX will not break
% a structure at a ~ so this keeps an author's name from being broken across
% two lines.
% use \thanks{} to gain access to the first footnote area
% a separate \thanks must be used for each paragraph as LaTeX2e's \thanks
% was not built to handle multiple paragraphs
%
\title{Parameter Efficient Multi-Class Intelligent Scheduling for Multimodal Online Distributed Industrial Anomaly Detection}

\author{{Heqiang Wang, Weihong Yang, Zheyuan Yang, Jia Zhou, Xiaoxiong Zhong, Fangming Liu, Weizhe Zhang} 
\thanks{
H. Wang, W. Yang, J. Zhou, X. Zhong, F. Liu and W. Zhang are with Pengcheng Laboratory, Shenzhen, 518066, China. Z. Yang is with Shenzhen International Graduate School, Tsinghua University, Shenzhen, 518055, China. (Corresponding Authors: Jia Zhou, Xiaoxiong Zhong.)}
} 

\maketitle

\begin{abstract}
Industrial anomaly detection has attracted significant attention as a fundamental challenge in industrial systems. The rapid advancement of heterogeneous industrial sensors has driven industrial anomaly detection from unimodal to multimodal paradigms. However, existing methods are primarily designed for centralized and offline settings, overlooking the distributed and continuously generated data characteristic of real-world industrial environments. With the advancement of edge intelligence, modern edge devices are increasingly capable of not only data acquisition but also distributed model training, enabling collaborative intelligence across the system. Industrial anomaly detection represents a critical application in this context. Motivated by these challenges, we propose a novel framework termed Multimodal Online Distributed Industrial Anomaly Detection (MODIAD). We first present a comprehensive workflow for MODIAD and then formulate a Multi-class Intelligent Scheduling (MIS) problem to coordinate cross class model updates by balancing data sufficiency and class update frequency. To efficiently solve this problem, we design a Sequential Marginal Gain Greedy (SMG) algorithm that enables effective multi-class training under resource constraints. Furthermore, to improve the computational and communication efficiency during training, we propose an Resource Efficient Class-Wise Low Rank Adaptation (REC-LoRA) strategy, which significantly reduces system overhead while preserving detection performance. Extensive experiments on two representative multimodal industrial anomaly detection datasets, MVTec 3D-AD and Eyecandies demonstrate that the proposed approach achieves superior performance and efficiency under the MODIAD scenario.
\end{abstract}

\keywords{Anomaly Detection, Distributed Learning, Multimodal Learning, Online Learning.}
% Note that keywords are not normally used for peerreview papers.
%\begin{IEEEkeywords}
%IEEEtran, journal, \LaTeX, paper, template.
%\end{IEEEkeywords}

% For peer review papers, you can put extra information on the cover
% page as needed:
% \ifCLASSOPTIONpeerreview
% \begin{center} \bfseries EDICS Category: 3-BBND \end{center}
% \fi
%
% For peerreview papers, this IEEEtran command inserts a page break and
% creates the second title. It will be ignored for other modes.
\IEEEpeerreviewmaketitle

\section{Introduction}
Industrial anomaly detection aims to identify defects or abnormal patterns in industrial products and plays a crucial role in automated quality inspection systems \cite{liu2024deep}. In practical industrial environments, defective samples are often scarce and difficult to obtain, making supervised learning approaches impractical. Consequently, most existing methods adopt an unsupervised setting, where models are trained solely on normal samples and detect anomalies during inference. Early research in industrial anomaly detection primarily focuses on single-modality data, particularly RGB images \cite{bergmann2019mvtec}. Although these approaches have achieved promising results, they often struggle in complex scenarios where visual appearance alone is insufficient. For instance, defects related to structural changes rather than color variations may be difficult to detect from 2D images, especially under varying lighting conditions \cite{cheng2026comprehensive}. To overcome these limitations, recent studies have shifted toward multimodal anomaly detection by integrating complementary data sources, such as RGB images and 3D point clouds \cite{wang2023multimodal}. Multimodal approaches enable the joint modeling of appearance and geometric information, thereby improving detection accuracy and robustness. Specifically, texture and color defects are more discernible in RGB images, while structural and shape-related anomalies are better captured by 3D data. Despite these advancements, most existing methods are developed under centralized and offline training frameworks, requiring all data to be collected and processed at a central server. This design limits scalability and adaptability in dynamic industrial environments where data is continuously generated in a distributed manner \cite{hamedi2025federated}.

\begin{figure}[htp]
\vspace{-5pt}
\centering
\subfloat{\includegraphics[width=0.95\linewidth]{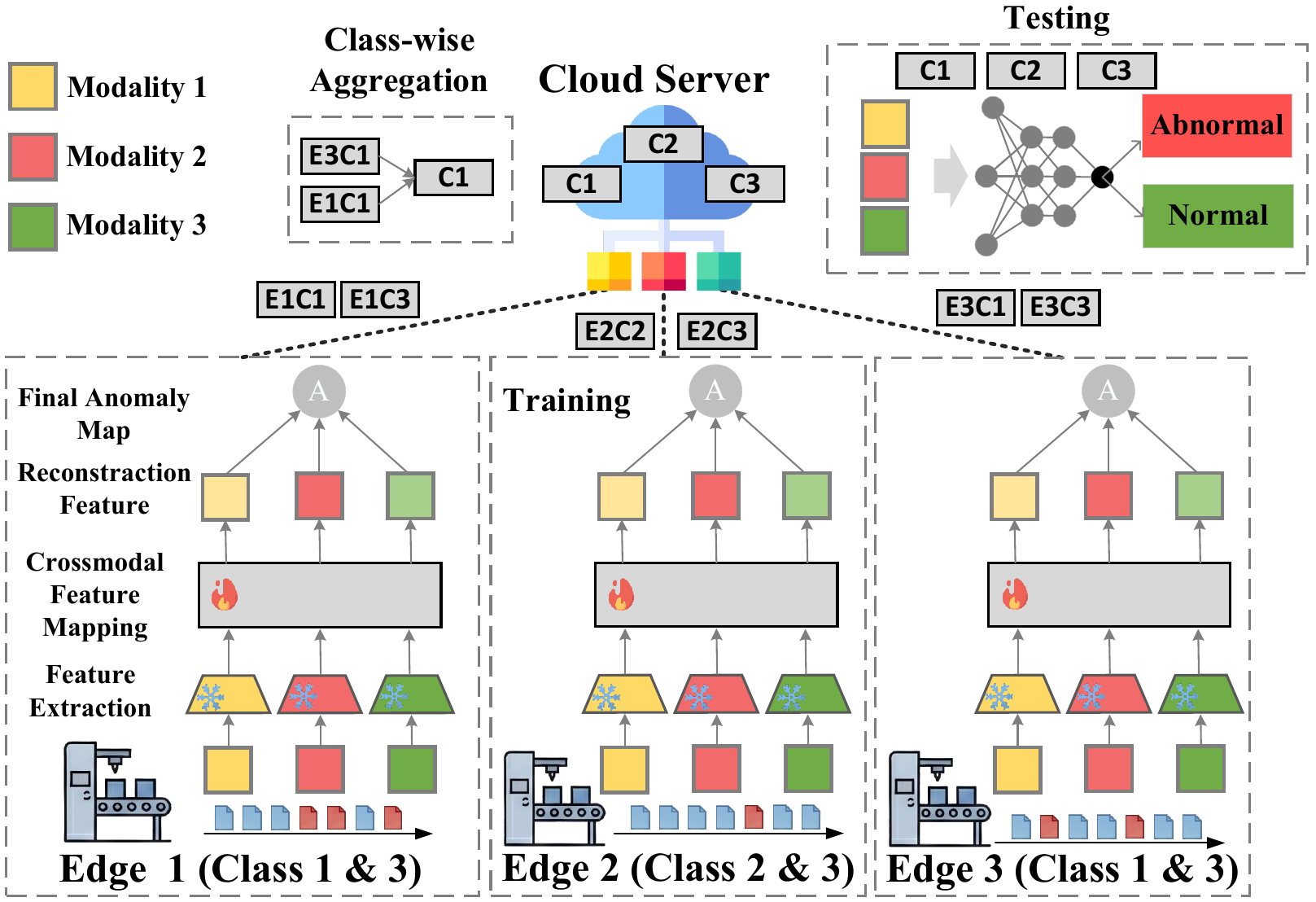}} 
\caption{Multimodal Online Distributed Industrial Anomaly Detection Framework (E1C1 denotes the local model corresponding to Class 1 on Edge Device 1.)}  
\label{modiad}
\vspace{-5pt}
\end{figure}

With the rapid advancement of edge intelligence devices in industrial environments, distributed learning has become increasingly feasible. In particular, federated learning (FL) \cite{zhang2021survey, xu2020client, xu2023federated} has emerged as an effective paradigm that enables multiple devices to collaboratively train models without sharing raw data, thereby reducing communication overhead and preserving data privacy. This framework provides a natural foundation for extending industrial anomaly detection from centralized settings to distributed environments, where data are generated and stored across multiple devices. However, compared to conventional FL tasks, distributed industrial anomaly detection introduces several unique challenges. First, the learning objective differs fundamentally. While FL typically focuses on supervised or semi-supervised tasks, industrial anomaly detection is predominantly unsupervised, aiming to model normal patterns and identify deviations. Second, most anomaly detection approaches adopt a one-for-one training strategy, in which a separate model is trained for each class. In resource-constrained edge environments, this design incurs substantial computational and communication overhead when multiple class-specific models must be updated. Consequently, unlike standard FL, distributed industrial anomaly detection requires explicit consideration of cross-class scheduling, determining which client–class pairs should be selected for training under limited system resources to achieve better overall detection performance. Moreover, the strong interdependence among these class-specific anomaly detection tasks fundamentally distinguishes this problem from conventional multi-task FL settings \cite{liu2022multi, zhou2022efficient}.

Despite recent advances in multimodal anomaly detection and distributed learning, existing approaches still face significant challenges in industrial deployments. In practice, data are naturally distributed across multiple edge devices such as sensors deployed along production lines, rather than being centrally aggregated. Moreover, industrial data are continuously generated in a streaming manner, which requires models to adapt to evolving data distributions instead of relying on static offline training. In addition, these data are inherently multimodal and involve heterogeneous sources such as RGB images, depth information, further increasing the complexity of detection system design. These characteristics limit the applicability of conventional FL frameworks, which are typically designed for static and single task scenarios. In particular, effectively coordinating distributed clients, handling streaming multimodal data, and supporting class-wise anomaly detection under constrained computational and communication resources remain open challenges. These limitations highlight the urgent need to develop a dedicated \textbf{Multimodal Online Distributed Industrial Anomaly Detection} (MODIAD) frameworks. To the best of our knowledge, this work is the first to investigate multimodal industrial anomaly detection in distributed and online learning settings. The main contributions of this work are summarized as follows:

\begin{enumerate}
    \item Motivated by the characteristics of industrial anomaly detection data, which is discretely distributed across devices, continuously generated in real time, and inherently multimodal, we introduce and formalize the concept of \textbf{Multimodal Online Distributed Industrial Anomaly Detection} (MODIAD). We further conduct a systematic study on how multiple edge intelligence devices can collaboratively perform anomaly detection under online multimodal data streams in IIoT environments.
    \item Given the limited computational and communication resources of edge intelligence devices in the MODIAD setting, our goal is to maximize training efficiency. To this end, we adopt a twofold strategy. First, we formulate and solve a Multi-class Intelligent Scheduling (MIS) problem that selects the most suitable client–class pairs in each round to maximize overall detection performance, thereby improving resource utilization. Second, we introduce an parameter-efficient strategy termed Resource-Efficien Class-Wise Low-Rank Adaptation (REC-LoRA) to further reduce computational and communication overhead during training.
    \item We evaluate the proposed strategies within the MODIAD framework on two representative multimodal industrial anomaly detection datasets, MVTec 3D-AD and Eyecandies. Experimental results demonstrate that the proposed approach achieves superior detection performance compared to baselines under the MODIAD setting. Furthermore, the results confirm that the proposed strategies effectively enhance computational and communication efficiency throughout the MODIAD training process.
\end{enumerate}

The remainder of this paper is organized as follows. Section II reviews related work on multimodal industrial anomaly detection, online federated learning, and communication and computation efficiency strategies. Section III presents the system model and formulates the problem. Section IV describes the workflow of the MODIAD framework. Section V introduces the proposed twofold strategy for improving computational and communication efficiency. Section VI reports experimental results that validate the effectiveness of the proposed approaches under the MODIAD setting. Finally, Section VII concludes the paper.

\section{Related Work }

\subsection{Multimodal Industrial Anomaly Detection}
Multimodal Industrial Anomaly Detection (MIAD) \cite{wang2023multimodal} extends traditional anomaly detection by leveraging heterogeneous sensor data commonly available in industrial environments. By integrating complementary modalities, such as visual and geometric information, MIAD improves both detection accuracy and robustness. The MVTec 3D-AD dataset \cite{bergmann2021mvtec} serves as a representative benchmark, providing paired RGB images and corresponding 3D point cloud data for each sample. Existing MIAD methods can be broadly categorized into three paradigms: (1) teacher–student approaches \cite{rudolph2023asymmetric, gu2024rethinking}, (2) memory bank based methods \cite{wang2023multimodal, cao2024complementary, tu2024self, chu2023shape}, and (3) reconstruction based methods \cite{chen2023easynet, costanzino2024multimodal}. Teacher–student methods, such as MMRD \cite{gu2024rethinking} and AST \cite{rudolph2023asymmetric}, learn modality-specific representations through knowledge distillation, enhancing anomaly detection by amplifying discrepancies between normal and abnormal samples. Memory bank–based methods, including M3DM \cite{wang2023multimodal} and Shape-Guided \cite{chu2023shape}, store multimodal feature representations and employ mechanisms such as contrastive learning and cross-modal alignment to improve detection and localization performance. Reconstruction-based methods, such as EasyNet \cite{chen2023easynet} and CFM \cite{costanzino2024multimodal}, model normal patterns by reconstructing multimodal inputs and identifying anomalies via reconstruction errors, often incorporating lightweight cross-modal mapping or adaptive fusion strategies. Despite their effectiveness, existing methods are primarily designed for centralized and offline settings, relying on pre-collected static datasets. This limitation hinders their applicability in real-world industrial environments, where data is continuously generated in a distributed manner and exhibits dynamic and heterogeneous characteristics. Addressing these challenges requires moving beyond conventional centralized and offline learning paradigms toward distributed and online multimodal anomaly detection frameworks. To the best of our knowledge, this work is the first to introduce the concept of multimodal online distributed industrial anomaly detection, along with a comprehensive workflow and corresponding algorithmic design.

\subsection{Online Federated Learning }
Online learning is designed to process data sequentially and update models incrementally, making it well-suited for applications involving continuously arriving data and the need for real-time model adaptation \cite{sahoo2017online}. These methods offer computational efficiency and eliminate the requirement of having access to the full dataset in advance, rendering them particularly suitable for memory-constrained IoT environments. In the context of FL, online federated learning (OFL) has emerged as a promising paradigm that extends online learning principles to distributed networks of decentralized learners \cite{hong2021communication}. A distinguishing feature of OFL, compared to traditional offline FL, is its emphasis on minimizing long-term cumulative regret rather than static optimization objectives during local updates. Although OFL remains relatively underexplored, several notable studies have recently advanced the field. For instance, \cite{kwon2023tighter} proposes a communication-efficient OFL algorithm that balances reduced communication overhead with strong learning performance. Similarly, \cite{mitra2021online} introduces FedOMD, an OFL method designed for uncertain environments, capable of handling streaming data without relying on assumptions about loss function distributions. While these works focus primarily on the horizontal federated learning (HFL) setting, \cite{wang2023online} explores the vertical federated learning (VFL) context \cite{wang2025pravfed}, proposing an online VFL framework tailored to cooperative spectrum sensing and achieving sublinear regret. Further extending to real-world industrial applications, \cite{wang2025denoising} addresses challenges such as noise interference and device heterogeneity in online VFL systems. In addition to the unimodal online federated learning approaches discussed above, recent studies have increasingly explored the role of multimodal data. For instance, \cite{wang2025multimodal} investigates the impact of missing modalities in multimodal online federated learning and proposes a corresponding compensation strategy. Meanwhile, \cite{wang2025mitigating} systematically examines the effects of data quantity and quality imbalance on learning performance and mitigates these issues through a rebalancing mechanism. Existing studies on MMO-FL are predominantly built upon FL frameworks and primarily focus on supervised tasks, relying on explicit label information and well-defined training objectives. In contrast, industrial anomaly detection is inherently an unsupervised problem, where the goal is to model normal patterns and identify deviations. As a result, it differs fundamentally from conventional MMO-FL in terms of modeling approaches, optimization objectives, and evaluation criteria, making existing methods difficult to directly apply.

\subsection{Communication and Computation Efficiency Strategy}
Communication and computation efficiency has become a fundamental challenge in edge intelligence and distributed learning systems, particularly in FL deployed over resource-constrained IoT devices \cite{almanifi2023communication}. FL introduces substantial overhead due to frequent communication rounds and local model updates, which are further exacerbated by limited bandwidth and device heterogeneity \cite{xu2020client, pfeiffer2023federated}. Notably, the strong coupling between communication and computation necessitates joint optimization rather than independent design \cite{wang2024efficient}. To address these challenges, existing approaches primarily focus on two directions. At the system level, client selection \cite{fu2023client} and task partitioning strategies \cite{mills2021multi} limit the number of participating devices or assign partial training tasks, thereby reducing communication cost and computational burden. At the model level, compression techniques such as pruning \cite{jiang2023computation}, quantization \cite{zhao2023aquila}, and parameter-efficient learning \cite{wang2025parameter} have been widely explored. In particular, Low-Rank Adaptation (LoRA) decomposes weight updates into low-rank structures, significantly reducing the number of trainable parameters and communication overhead while preserving model expressiveness \cite{liu2025adaptive}. However, most existing methods consider client-level selection and model-level compression separately, without fully leveraging their joint potential. This limitation becomes more pronounced in multimodal distributed industrial anomaly detection, where class heterogeneity across clients further complicates resource allocation. To address these issues, we propose a unified framework that jointly optimizes task scheduling and model adaptation. The proposed approach effectively reduces both computational and communication costs while preserving sufficient model capacity for classes with complex anomaly characteristics, thereby providing an efficient solution for distributed anomaly detection.

\section{System Model}
To illustrate the MODIAD process, we consider a smart factory scenario consisting of a central cloud server and $K$ edge intelligence clients that jointly provide anomaly detection services for the production line. Each client processes online multimodal data collected from heterogeneous sensors, such as RGB images captured by cameras and point cloud data acquired from 3D sensors. Owing to improved on-device computational capabilities, communication efficiency considerations, and privacy requirements, raw data is retained locally at the edge rather than being uploaded to the cloud. Through local model training, each client learns the representations of normal operational patterns and periodically transmits model updates to the central cloud server for aggregation, thereby enabling collaborative anomaly detection. 

During production line operation, each client continuously collects new data over time, and the overall timeline is partitioned into discrete intervals indexed by $t = 1, 2, ..., T$. Each interval is treated as a global round in MODIAD process. In each global round $t$, each client $k \in \mathcal{K}$ collects the current local training dataset $\mathcal{D}_k^t$, which consists of normal multimodal samples spanning $M$ modalities and $C$ object classes. Since industrial anomaly detection usually learns from defect-free data, $\mathcal{D}_k^t$ contains only normal samples collected from the local production environment. These samples may belong to different object classes and contain multiple sensing modalities. The goal is to collaboratively learn a global anomaly detection model without sharing raw data. Formally, the local dataset of client $k$ at round $t$ is decomposed into class-wise subsets as:
\begin{align}
\mathcal{D}_k^t = \left ( \mathcal{D}_{k, 1}^{t}, \dots, \mathcal{D}_{k, c}^{t}, \dots, \mathcal{D}_{k, C}^{t} \right )
\end{align}
where $\mathcal{D}_{k, c}^{t}$ denotes the subset of normal samples belonging to class $c$  at client $k$. Due to the Non-IID nature of distributed industrial environments, a client may only observe a subset of all classes at a given round $t$. Therefore, we define the locally available class set of client $k$ at round $t$ as $\mathcal{I}_k^t = \left\{c \mid \mathcal{D}_{k,c}^t \neq \emptyset \right\}$. However, due to computational and communication constraints, not every class $c$ owned by client $k$ will participate in training. Each class-wise subset is further organized according to sensing modalities. Suppose there are $M$ modalities. Then the class-wise dataset $\mathcal{D}_{k, c}^{t}$ can be represented as:
\begin{align}
\mathcal{D}_{k, c}^{t} = \left ( X_{k, c, 1}^{t}, X_{k, c, 2}^{t}, \dots, X_{k, c, m}^{t}, \dots, X_{k, c, M}^{t} \right ) 
\end{align}
while $X_{k, c, m}^{t}$ the set of modality $m$ samples of class $c$ collected by client $k$ at round $t$. For convenience, a multimodal sample is denoted as $x =\left \{x_1,x_2,\dots,x_M\right\}$. Without loss of generality, we assume that each collected sample contains complete modality information. The overall distributed training data at round t, together with its class-wise decomposition, can be conceptually expressed as:
\begin{align}
\mathcal{D}^t = \bigcup_{k=1}^{K} \mathcal{D}_k^t, \quad\mathcal{D}^t_c = \bigcup_{k=1}^{K} \mathcal{D}_{k, c}^t
\end{align}
It should be noted that $\mathcal{D}^t$ is only a conceptual global dataset and is never physically centralized at the server.

To obtain discriminative representations for industrial anomaly detection, each modality is processed by a corresponding frozen pre-trained feature extractor. For modality $m$, we denote the frozen feature extractor as $\Phi^{frz}_m$. Given a multimodal sample $x$, the extracted representation of modality $m$ is expressed as $E_m = \Phi^{frz}_m(x_m)$. The multimodal feature representation can then be written as:
\begin{align}
E = \Phi^{\mathrm{frz}}(x) = \left\{ \Phi_1^{\mathrm{frz}}(x_1), \dots, \Phi_M^{\mathrm{frz}}(x_M) \right\}
\end{align}

These frozen modality-specific feature extractors reduce the computational burden of local training and allow the subsequent learning process to focus on modeling normal cross-modal relationships. This design is particularly suitable for distributed industrial environments, where edge devices usually have limited computational resources. In this work, the feature extractors remain fixed during distributed training, and only the downstream mapping models are updated.

Different from conventional classification-oriented FL, the goal of industrial anomaly detection is to learn the feature distribution of normal samples and identify deviations from this distribution during inference. Moreover, since normal appearance and geometric structures vary significantly across different industrial object classes, a one-for-one modeling strategy is adopted. Under this strategy, each class is associated with an independent feature mapping reconstruction model. The set of class-specific mapping models is denoted as
\begin{align}
\Theta = \left\{\Theta_1,  \dots, \Theta_c, \dots, \Theta_C \right\} 
\end{align}
where $\Theta_c$ represents the feature mapping model for class $c$. For a sample $x$ belonging to class $c$, the corresponding class-specific model maps its frozen multimodal feature representation to a reconstructed feature representation as $\hat{E} = \Theta_c(\Phi^{frz}(x))$. The discrepancy between the reconstructed feature $\hat{E}$ and the original feature $E$ is used to measure whether the sample follows the learned normal pattern. Accordingly, the learning objective of the class-specific anomaly detection model is formulated by minimizing the reconstruction error over normal samples. The overall MODIAD learning objective over all classes can be formulated as:
\begin{align}
\min_{\Theta} \mathcal{L}^{t} = \sum_{c=1}^{C} \frac{1}{|\mathcal{D}_c^t|} \sum_{x \in\mathcal{D}_c^t} \ell_{\mathrm{AD}} \left( \Theta_c\left( \Phi^{\mathrm{frz}}(x) \right),
\Phi^{\mathrm{frz}}(x)\right)
\end{align}
where $\ell_{\mathrm{AD}}$ denotes the anomaly detection loss. This objective indicates that MODIAD aims to learn a bank of class-specific anomaly detection models, where each model captures the normal multimodal feature distribution of its corresponding class. Since $\mathcal{D}_c^t$ is distributed across clients and cannot be centralized, this objective is optimized through local model updates at clients and class-wise aggregation at the server. The above formulation provides a general system-level abstraction of MODIAD, but does not specify how client-server interactions are performed in each global round. In the next section, we present the MODIAD workflow to highlight its key differences from the traditional FL process.

\section{The Workflow of MODIAD}
Based on the general system model defined in the previous section, we now describe the round-wise workflow of MODIAD. While the system model provides a high-level abstraction of the MODIAD problem, this section further instantiates the workflow through a cross-modal feature mapping mechanism. Specifically, we consider a multimodal industrial anomaly detection setting with 2D RGB images and 3D point clouds as representative modalities, and use this scenario to illustrate how class-specific anomaly detection models are trained and updated across clients.

Different from conventional federated learning, where each client typically updates a single global model, MODIAD adopts a one-for-one class-wise anomaly detection strategy. Therefore, each object class is associated with an independent anomaly detection model. In this work, each class-specific model is instantiated by a pair of crossmodal mapping networks. Specifically, for class $c$, we have:
\begin{align}
\Theta_c = \left\{M^{2D\rightarrow 3D}_{c}, M^{3D\rightarrow 2D}_{c} \right\}
\end{align}
where $M^{2D\rightarrow 3D}_{c}$ maps 2D features into the 3D feature space, and $M^{3D\rightarrow 2D}_{c}$ maps 3D features into the 2D feature space. This design is inspired by the general principle of cross-modal feature mapping \cite{costanzino2024multimodal}, where the relationships between features extracted from nominal samples are modeled, and anomalies are identified by measuring the inconsistency between observed features and their predicted cross-modal counterparts. In each global round $t \in \mathcal{T}$, where $\mathcal{T} = \{ 0, 1, 2, \dots, T-1 \}$, the training process proceeds as follows.

\subsection{Training Workflow}

\subsubsection{\textbf{Client - Online Data Collection}} At the beginning of each global round $t$, each client $k$ receives newly arrived multimodal normal samples from its local production environment and incorporates them into the local dataset $\mathcal{D}_k^t$. Since the data are organized by class, the client maintains class-wise subsets $\mathcal{D}_{k,c}^{t}$. For each class $c$, client $k$ records the number of locally available samples as $N_{k,c}^{t} = |\mathcal{D}_{k,c}^{t}|$. This statistic reflects the data sufficiency of class $c$ at client $k$. Since data arrive online and the distribution across clients is Non-IID, the available class set $\mathcal{I}_k^t $ may vary across clients and training rounds.

\subsubsection{\textbf{Client - Status Report Upload}} Instead of uploading raw multimodal data, each client sends lightweight status information to the server. The status report primarily includes the class-wise sample statistics $\left\{N_{k,c}^{t} \right\}_{c = 1}^C$ for the current round. Based on these local data distribution statistics, the server subsequently assigns anomaly detection training tasks to each client.

\subsubsection{\textbf{Server - Global Scheduling and Model Broadcast}} After receiving the status reports from all clients, the server determines which class-specific models should be updated in the current round. Let $\mathcal{C}_k^t \subseteq \mathcal{I}_k^t$ denote the set of classes assigned to client $k$ for updating at round $t$, where $\mathcal{I}_k^t$ represents the locally available classes at client $k$, and $\mathcal{C}_k^t$ denotes the subset selected by the server for training. The scheduling principles used to determine $\mathcal{C}_k^t$ will be detailed in the next section. For each scheduled class $c\in\mathcal{C}_k^t$, the server broadcasts the corresponding global class-specific cross-modal
feature mapping model $\Theta_c^t$ to client $k$, which serves as the initialization for the new local training round.

\subsubsection{\textbf{Client - Multimodal Feature Extraction}} For each scheduled class $c\in\mathcal{C}_k^t$, client $k$ processes the corresponding local normal samples in $\mathcal{D}_{k,c}^{t}$. Specifically, the client first extracts 2D and 3D feature representations using frozen modality-specific feature extractors:
\begin{align}
E_{2D} = \Phi^{\mathrm{frz}}_{2D}(x_{2D}), \quad E_{3D} = \Phi^{\mathrm{frz}}_{3D}(x_{3D})
\end{align}
Here we use frozen 2D and 3D feature extractors and keeps their weights fixed during training. The extracted 2D and 3D feature maps are aligned at the pixel level so that features from both modalities can be compared spatially.

\subsubsection{\textbf{Client -  Crossmodal Mapping Model Update}} For each scheduled class $c$, client $k$ jointly trains two class-specific cross-modal mapping networks. Given the aligned feature maps $E_{2D}$ and $E_{3D}$, the two mapping networks generate crossmodal feature predictions as:
\begin{align}
\hat{E}_{3D} = M^{2D\rightarrow 3D}_{c} ({E}_{2D}), \quad \hat{E}_{2D} = M^{3D\rightarrow 2D}_{c} ({E}_{3D})
\end{align}
The objective is to learn the relationship between the 2D and 3D feature spaces using nominal samples. Specifically, the local training objective minimizes the discrepancy between the observed features and their predicted cross-modal counterparts. The local loss is defined as:
\begin{align}
\mathcal{L}_{k,c}^{t} = \frac{1}{|\mathcal{D}_{k,c}^{t}|} \sum_{x\in\mathcal{D}_{k,c}^{t}}\left[d_{\cos} \left(E_{2D}, \hat{E}_{2D}\right)
+d_{\cos}\left(E_{3D},\hat{E}_{3D}\right)\right]
\end{align}
where $d_{\cos}(\cdot,\cdot)$ denotes the cosine distance in the feature space. During local training, client $k$ performs $\tau_{max}$ local iterations for each scheduled class. The update process can be written as:
 \begin{align}
        &\Theta_{k, c}^{t,0} = \Theta^{t}_c, \notag\\
        &\Theta_{k, c}^{t, \tau + 1} = \Theta_{k, c}^{t, \tau } - \eta \textbf{G}_{k, c}^{t, \tau }, \quad \forall \tau = 1, ..., \tau_{max} \notag\\
        &\Theta_{k, c}^{t+1} = \Theta_{k, c}^{t, \tau_{max}} 
\end{align}
where $\textbf{G}_{k, c}^{t, \tau }$ denotes the gradient computed from the local loss, and $\eta$ is the learning rate. Since the feature extractors are frozen, only the class-specific mapping networks are updated during local training.

\subsubsection{\textbf{Client - Local Model Upload}} After completing local training, client $k$ uploads only the updated models corresponding to the scheduled class set $\mathcal{C}_k^t$ to the central server for aggregation as: $\Theta_k^{t+1}= \left\{\Theta_{k,c}^{t+1}\right\}_{c\in\mathcal{C}_k^t}$. This selective upload strategy prioritizes the client–class pairs that contribute most to improving overall anomaly detection performance, while simultaneously reducing computational and communication overhead.

\subsubsection{\textbf{Server - Class-wise Global Model Aggregation}} After receiving local model updates, the server aggregates the uploaded models independently for each class. For class $c$, let $\mathcal{P}^t_c =  \left\{k\mid c\in\mathcal{C}_k^t\right\}$ denote the set of clients that uploaded the corresponding class $c$ model in round $t$. The server updates the global class-specific model by:
\begin{align}
    \Theta^{t+1}_c = \frac{1}{|\mathcal{P}^t_c|}\sum_{k \in \mathcal{P}^t_c} \Theta^{t+1}_{k,c}
\end{align}
If no client uploads an update for class $c$ in round $t$, the server keeps the previous global model unchanged. After completing aggregation for all classes, the server obtains the updated global model bank as:
\begin{align}
\Theta^{t+1} = \left\{ \Theta_1^{t+1}, \Theta_2^{t+1}, \dots, \Theta_C^{t+1}\right\}
\end{align}
This model bank is stored at the server and used for initialization in the next global round.

\subsection{Testing Workflow}
After training, the learned global class-specific model bank is used for anomaly detection to evaluate the learning performance. Given a test sample, its multimodal feature representations are first extracted using the frozen feature extractors, and the corresponding cross-modal predictions are then generated by the class-specific mapping networks. Based on the discrepancy between the observed features and the predicted cross-modal features, the modality-specific anomaly maps can be obtained as follows:
\begin{align}
\Psi_{2D} = \textit{Disc} \left ( E_{2D}, \hat{E}_{2D} \right ), \Psi_{3D} = \textit{Disc} \left ( E_{3D}, \hat{E}_{3D} \right )
\end{align}
where $\textit{Disc}(\cdot,\cdot)$ denotes a feature discrepancy function, such as Euclidean distance after normalization. The two modality-specific anomaly maps are then fused to obtain the final anomaly map. In this work, we adopt pixel-wise product aggregation:
\begin{align}
\Psi = \Psi_{2D} \cdot \Psi_{3D}
\end{align}
This aggregation can be interpreted as a logical AND operation, where a spatial location receives a high anomaly score only when both modality-specific discrepancies are large. 

From the above workflow, MODIAD can be seen to differ fundamentally from conventional online FL. While traditional FL typically requires each client to update a single global model in each round, MODIAD adopts a one-for-one anomaly detection strategy, where each object class corresponds to an independent class-specific model. As a result, a client may have multiple available class-wise training tasks in the same round, but limited computational and communication resources make it impractical to update all of them simultaneously. This raises a key decision-making challenge: how to select appropriate client–class pairs for training at each global round. An effective strategy should consider not only local data sufficiency, but also global class balance, so that all classes receive adequate training over time. Otherwise, frequently updated classes may dominate the learning process, while under trained classes degrade overall anomaly detection performance. To address this challenge, we formulate the Multi-class Intelligent Scheduling (MIS) problem, which determines which clients should update which class-specific models in each round under resource constraints, thereby improving the overall effectiveness of MODIAD in distributed online learning settings.

\section{Multi-class Intelligent Scheduling Problem}
Following the motivation above, MIS is introduced to coordinate class-wise model updates across clients under limited resources. The key issue is how to quantify the update priority of each candidate client–class pair. In MODIAD, this priority should jointly capture local and global factors. From the local perspective, a client–class pair is meaningful for training only when sufficient normal samples have been accumulated to support reliable model updates. From the global perspective, the scheduler should avoid class-wise update imbalance, since insufficiently updated class-specific models may degrade the overall anomaly detection performance. Therefore, we define two scheduling principles, namely data sufficiency and global class balance, and use them to construct the priority score for client–class selection.

\begin{table*}[t]
\centering
\caption{I-AUROC and AUPRO@10\% on MVTec 3D-AD for Various Data Proportions. Best results in \textbf{bold}. (Epoch@10)}
\label{tab:sufficient}
\resizebox{\textwidth}{!}{
\begin{tabular}{c|c|cccccccccc|c}
\hline
\multirow{2}{*}{Metric} & \multirow{2}{*}{Proportion} 
& \multicolumn{10}{c|}{Class} & \multirow{2}{*}{Mean} \\
\cline{3-12}
 &  & Bagel & Cable & Carrot & Cookie & Dowel & Foam & Peach & Potato & Rope & Tire &  \\
\hline
\multirow{3}{*}{I-AUROC}
& 0.2     & 0.964 & 0.704 & 0.945 & 0.984 & 0.864 & 0.746 & 0.867 & 0.783 & 0.944 & 0.595 & 0.840 \\
& 0.5    & 0.978 & \textbf{0.776} & 0.953 & 0.992 & 0.933 & 0.768 & 0.929 & 0.817 & 0.946 & 0.701 & 0.879 \\
& 0.8  & \textbf{0.980} & 0.769 & \textbf{0.971} & \textbf{0.993} & \textbf{0.938} & \textbf{0.796} & \textbf{0.942} & \textbf{0.844} & \textbf{0.955} & \textbf{0.736} & \textbf{0.892} \\
\hline

\multirow{3}{*}{AUPRO@10\%}
& 0.2   & 0.929 & 0.583 & 0.942 & 0.889 & 0.733 & 0.785 & 0.915 & 0.936 & 0.907 & 0.827 & 0.845 \\
& 0.5   & \textbf{0.935} & 0.738 & 0.944 & 0.892 & 0.778 & 0.835 & 0.933 & 0.942 & 0.914 & 0.869 & 0.878 \\
& 0.8  & \textbf{0.935} & \textbf{0.796} & \textbf{0.946} & \textbf{0.893} & \textbf{0.797} & \textbf{0.853} & \textbf{0.936} & \textbf{0.945} & \textbf{0.916} & \textbf{0.896} & \textbf{0.891} \\
\hline
\end{tabular}
}
\end{table*}

\subsection{Scheduling Principles}

\textbf{(1) Data Sufficiency:} To improve learning efficiency, the system prioritizes classes with sufficient training data. Updating models for classes with insufficient data may lead to suboptimal learning outcomes while unnecessarily consuming computational and communication resources. To further investigate the impact of normal data volume on anomaly detection performance, we conduct a centralized evaluation on the MVTec 3D-AD dataset, examining how detection performance across different classes varies with data availability, as illustrated in Tab.~\ref{tab:sufficient}. As observed from this table, a larger proportion of available normal training data generally leads to improved detection performance across classes. Detailed experimental settings can be found in the experiment section below. Motivated by these observations, we define a data sufficiency score as follows:
\begin{align}
R_{k,c}^t = \log(1 + N_{k,c}^t)
\end{align}
where $N_{k,c}^t$ denotes the number of samples of class $c$ collected by client $k$ at round $t$. This formulation reflects the importance of sample quantity while preventing dominance by excessively large datasets.

\begin{table*}[t]
\centering
\caption{I-AUROC and AUPRO@10\% on MVTec 3D-AD for Various Training Epoch. Best results in \textbf{bold}. (Proportion@0.5)}
\label{tab:balanced}
\resizebox{\textwidth}{!}{
\begin{tabular}{c|c|cccccccccc|c}
\hline
\multirow{2}{*}{Metric} & \multirow{2}{*}{Epoch} 
& \multicolumn{10}{c|}{Class} & \multirow{2}{*}{Mean} \\
\cline{3-12}
 &  & Bagel & Cable & Carrot & Cookie & Dowel & Foam & Peach & Potato & Rope & Tire &  \\
\hline
\multirow{3}{*}{I-AUROC}
& 10   & 0.980 & 0.776 & 0.953 & 0.992 & 0.933 & 0.768 & 0.929 & 0.817 & 0.946 & 0.701 & 0.879 \\
& 20    & \textbf{0.986} & 0.787 & 0.974 & \textbf{0.994} & 0.951 & 0.843 & 0.941 & 0.866 & 0.961 & 0.774 & 0.908 \\
& 50   & 0.981 & \textbf{0.832} & \textbf{0.977} & \textbf{0.994} & \textbf{0.975} & \textbf{0.860} & \textbf{0.963} & \textbf{0.901} & \textbf{0.968} & \textbf{0.848} &\textbf{ 0.930} \\
\hline

\multirow{3}{*}{AUPRO@10\%}
& 10   & 0.935 & 0.738 & 0.944 & 0.892 & 0.778 & 0.835 & 0.932 & 0.942 & 0.914 & 0.869 & 0.878 \\
& 20    & 0.936 & 0.824 & 0.945 & \textbf{0.894} & 0.798 & 0.871 & \textbf{0.938} & 0.946 & 0.919 & 0.914 & 0.899 \\
& 50   & \textbf{0.938} & \textbf{0.874} & \textbf{0.947} & \textbf{0.894} & \textbf{0.835} & \textbf{0.875} & \textbf{0.938} & \textbf{0.948} & \textbf{0.923} & \textbf{0.933} & \textbf{0.911} \\
\hline
\end{tabular}
}
\end{table*}

\textbf{(2) Class Balance:} From a global perspective, it is important to ensure that all classes receive sufficient training updates. If certain classes are updated significantly more frequently than others, some class models may remain insufficiently trained. Such class imbalance can lead to undetected anomalies during inference and compromise the reliability of industrial anomaly detection systems. To further investigate the impact of cross-class balance on anomaly detection performance, we conduct a centralized evaluation on the MVTec 3D-AD dataset, analyzing how detection performance varies across classes under different numbers of epochs, as illustrated in Tab.~\ref{tab:balanced}. The results indicate that increasing the number of training epochs generally improves the anomaly detection performance of most classes, suggesting that class-specific anomaly detection models benefit from sufficient update opportunities. This observation further motivates the need for globally balanced class-wise scheduling. Firstly, we define the cumulative number of updates for class c up to round $t$ as:
\begin{align}
B_c^t =\sum_{t'=1}^{t} \sum_{k=1}^{K} b_{k,c}^{t'}
\end{align}
where $b_{k,c}^{t'} \in  \left\{0, 1 \right\}$ is a binary indicator representing whether client $k$ updates the model of class $c$ in round $t'$. Based on this definition, the global class update distribution at round $t$ can be expressed as:
\begin{align}
\pi_c^t =\frac{B_c^t}{\sum_{c'=1}^{C} B_{c'}^t}
\end{align}
Ideally, all classes should receive balanced update opportunities over time. Finally, the class imbalance weight is defined as
\begin{align}
V^t_c = \log  \left ( 1+\frac{1}{\pi_c^t +\epsilon}\right )
\end{align}
where $\epsilon$ is a small constant introduced to ensure numerical stability. This weight assigns higher priority to classes with smaller update proportions, encouraging the scheduler to compensate for under-updated classes and maintain a more balanced class-wise training process.  

Based on the above two principles, the server computes a priority score for each candidate client-class pair. The priority metric for selecting class-specific updates is defined as:
\begin{align} 
 U_{k,c}^t &= \alpha \bar{R}_{k,c}^t + \beta  \bar{V}_c^t \notag \\
&= \alpha \frac{R_{k,c}^t}{\max_{k,c} R_{k,c}^t}
+ \beta \frac{V_c^t}{\sum_{c'=1}^{C} V_{c'}^t}
\end{align}
Where $\alpha$ and $\beta$ are weighting hyperparameters that control the relative contributions of data sufficiency and class balance, respectively. Based on this priority metric, the server determines the class update schedule by solving the following optimization problem:
\begin{align}
& \max_{b_{k,c}^{t}} \sum_{k=1}^K \sum_{c=1}^C b_{k,c}^{t}  U_{k,c}^t \\
&  s.t. \quad b_{k,c}^{t} \in  \left\{0, 1 \right\} \\
& \quad \quad \sum_{c=1}^C b_{k,c}^{t} \leq \Gamma^{cop}_k \label{cons1}\\
& \quad \quad  \sum_{k=1}^K \sum_{c=1}^C b_{k,c}^{t} \leq \Gamma^{com} \label{cons2}
\end{align}
where $\Gamma^{cop}_k$ denotes the maximum number of class models that client $k$ can update in every round, reflecting its local computational capacity, and $\Gamma^{com}$ denotes the maximum number of model updates that the system can accommodate per round, constrained by the communication bandwidth between clients and the server. In the next section, we will provide a detailed solution algorithm to the above problem.

\subsection{Sequential Marginal Gain Greedy Solver} Since decisions made in the current round influence subsequent outcomes, the MIS problem becomes a decision-dependent and coupled 0–1 optimization problem. To address this challenge, we adopt a Sequential Marginal Gain Greedy (SMG) solver, where the server sequentially selects client-class pairs according to their marginal contributions. Let $\mathcal{S}\subseteq\{(k,c)\}$ denote the set of selected client-class pairs in the current round. Given $\mathcal{S}$, the cumulative update count of class $c$ can be updated as:
\begin{align} 
B_c^{t}(\mathcal{S}) = B_c^{t-1} +  \kappa _c(\mathcal{S}),
\end{align}
where $ \kappa _c(\mathcal{S}) = \sum_{k=1}^K b_{k,c}^{t}(\mathcal{S})$ denotes the number of selected updates for class $c$ in the current round. Then for any feasible candidate pair $(k,c)\notin \mathcal{S}$, the marginal gain is defined as:
\begin{align} 
& F(\mathcal{S}) =
\sum_{(k,c)\in \mathcal{S}}\left(\alpha \bar{R}_{k,c}^{t}+\beta \bar{V}_c(\mathcal{S})\right)\notag  \\
& \Delta_{k,c}(\mathcal{S}) =  F(\mathcal{S} \cup {(k,c)}) - F(\mathcal{S})
\end{align}
where $\bar{V}_c(\mathcal{S})$ is computed using the updated $B_c^{t}(\mathcal{S})$. This marginal gain captures the local benefit from selecting $(k,c)$ and global effect on class imbalance through $\bar{V}_c$. At each greedy step, the server selects the feasible candidate with the largest marginal gain:
\begin{align} 
&(k^\star,c^\star)=\arg\max_{(k,c)\in\Omega(\mathcal{S})}\Delta_{k,c}(\mathcal{S}) 
\end{align}
where the feasible candidate set is defined as:
\begin{align}
\Omega(\mathcal{S})=\{(k,c)\mid(k,c)\notin\mathcal{S}, \mathrm{~} c \in \mathcal{I}_k^t, \mathrm{~}Eq.~\ref{cons1},\mathrm{~}Eq.~\ref{cons2}\} \notag 
\end{align}
The selected pair is added into $\mathcal{S}$, and and the corresponding class-wise statistics are updated accordingly. This process is repeated until the stopping condition is met, such as reaching the communication budget, exhausting all feasible candidates, or obtaining no positive marginal gain. Finally, the scheduled class set for client \(k\) is obtained as $\mathcal{C}_k^t = \{c\mid (k,c)\in\mathcal{S}\}$. Each client then updates the assigned class-specific models according to \(\mathcal{C}_k^t\) and proceeds with the subsequent workflow.

\subsection{Resource Efficient Class-Wise Low Rank Adaptation}
The preceding scheduling step reduces the system-level overhead of MODIAD by selecting only a subset of client–class pairs for model updates in each global round. However, even after scheduling, updating and uploading multiple class-specific mapping models may still incur considerable computational and communication costs, especially when the number of classes is large. To further improve training efficiency, we introduce a Resource Efficient Class-Wise Low Rank Adaptation (REC-LoRA) strategy. Unlike the scheduling step, which determines which client–class pairs should be updated, this strategy determines how each selected class-specific model should be updated.

The core idea is to adapt the update mechanism according to the learning status of each class. Since different industrial object classes may exhibit heterogeneous convergence behaviors, applying a uniform update strategy to all classes can be inefficient. For classes whose anomaly detection performance has become relatively stable, parameter efficient low rank updates are sufficient to maintain and refine the model. In contrast, for classes whose performance remains unsatisfactory, full model updates are preserved to provide stronger learning capacity. Accordingly, the training process is divided into two phases.

\textbf{Warm up Phase:} In the early stage of training, each class specific crossmodal mapping model undergoes a warm up phase of $T_{\mathrm{warm}}$ rounds. During this phase, full model updates are performed for the corresponding bidirectional crossmodal mapping networks. This phase enables each class specific model to learn sufficiently expressive normal 2D-3D feature relationships and provides a reliable initialization for subsequent low-rank adaptation.

\textbf{Adaptive Hybrid Update Phase:} After the warm-up phase, the update strategy for each class is adaptively determined based on its relative detection performance. Let $Q_c^{t-1}$ denote the validation performance of class $c$ at round $t-1$, measured by detection metrics such as image-level AUROC or AUPRO. To reduce the influence of fluctuations caused by online training, we maintain a smoothed class-wise performance score:
\begin{align}
\bar Q_c^{t-1} = \gamma Q_c^{t-1} + (1-\gamma)\bar Q_c^{t-2}, \quad \gamma\in(0,1]
\end{align}
where $\gamma$ controls the degree of temporal smoothing.The average smoothed detection performance across all classes is defined as:
\begin{align}
Q_{\mathrm{avg}}^{t-1} = \frac{1}{C} \sum_{c=1}^{C} \bar Q_c^{t-1}
\end{align}
Based on the relative class-wise performance, we define the update-mode indicator as:
\begin{align} 
z_c^t =
\begin{cases}
1, & \bar Q_c^{t-1} \ge Q_{\mathrm{avg}}^{t-1}  \\
0, & \text{otherwise}
\end{cases}
\end{align}
When $z_c^t = 1$ , the anomaly detection performance for class $c$ is considered relatively satisfactory, and a LoRA-based update can be adopted to reduce computational and communication overhead. Conversely, when $z_c^t = 0$, the detection performance has not yet reached a convincing level, and full model updates are retained to ensure adequate learning capacity.

For a class selected by the MIS scheduler, the actual update form is determined by $z_c^t$. If $z_c^t = 0$, all trainable parameters of the corresponding class-specific mapping model are updated. If $z_c^t = 1$, the base weights are frozen and only lightweight low-rank residual parameters are optimized and uploaded.

The basic idea of LoRA is to avoid directly updating a full weight matrix. Instead, it keeps the original weight fixed and learns a low-rank residual update.  For the $l$-th linear layer of the class-specific mapping network for class \(c\), let the frozen base weight be:
\begin{align}
W_{l,c}^{\mathrm{base}} \in \mathbb{R}^{d_{l}^{out}\times d_{l}^{in}}
\end{align}
When a class switches to LoRA-based updating, the most recent full-model weight is treated as the frozen base weight. The effective weight used during forward propagation is written as:
\begin{align}
\widetilde W_{l,c}^{t} = W_{l,c}^{\mathrm{base}} + \Delta W_{l,c}^{t},
\end{align}
where the residual update is parameterized by a low-rank decomposition:
\begin{align}
& \Delta W_{l,c}^{t} = H_{l,c}^{t} J_{l,c}^{t} \\ 
& H_{l,c}^{t} \in \mathbb{R}^{d_{l}^{out}\times r_c^t},
J_{l,c}^{t} \in \mathbb{R}^{r_c^t\times d_{l}^{in}},
\end{align}
where $r_c^t\ll \min(d_{l}^{in},d_{l}^{out})$ denotes the low-rank dimension assigned to class $c$ at round $t$. Therefore, instead of learning a full update matrix with $d_{l}^{out}d_{l}^{in}$ parameters, REC-LoRA only learns $r_c^t(d_{l}^{out}+d_{l}^{in})$ trainable parameters for this layer. This substantially reduces the number of parameters involved in local update and distributed upload.

\section{Experiment}
In this section, we experimentally evaluate the performance of the proposed MODIAD algorithm under the considered setting. All experiments are conducted on a workstation running Ubuntu 24.04, equipped with an Intel Core i9-14900K CPU and an NVIDIA GeForce RTX 4090 GPU. The detailed experimental settings are described as follows.

\subsection{Datasets}
\textbf{MVTec 3D-AD}: The MVTec 3D-AD dataset is a widely used benchmark for multimodal industrial anomaly detection, designed to evaluate algorithms under realistic manufacturing conditions where defects may occur in both visual appearance and geometric structure. The dataset consists of 10 industrial object categories, with 2,656 training samples, 294 validation samples, and 1,197 test samples. The training set is composed exclusively of defect-free samples, whereas the test set contains both normal and anomalous samples. The anomalies include scratches, dents, contamination, and structural deformation. Data is captured using a structured-light 3D sensor, which simultaneously acquires high-resolution RGB images and depth information. Each sample therefore contains two complementary modalities: appearance and geometry, making the dataset well suited for evaluating multimodal anomaly detection methods in industrial environments.

\textbf{Eyescandies}: The Eyecandies dataset \cite{bonfiglioli2022eyecandies} is a synthetic benchmark consisting of photorealistic images of 10 categories of food items captured in an industrial conveyor scenario. It contains 10,000 training samples, 1,000 validation samples, and 4,000 test samples. Each category includes multiple defect types that simulate common imperfections in industrial production. The training set consists exclusively of defect-free samples, while the test set includes both normal and anomalous instances. Data is acquired using a multimodal sensing setup that simultaneously captures RGB images and three-dimensional depth information, providing complementary appearance and geometric cues. In our experiments, the Eyecandies dataset is used as a supplementary benchmark to validate the results obtained on the MVTec 3D-AD dataset.

\subsection{Evaluation Metrics} Unlike traditional FL, which commonly uses test accuracy as the evaluation metric, we evaluate the anomaly detection performance using the Area Under the Receiver Operating Characteristic curve at the image level (I-AUROC), computed based on the global anomaly scores. For anomaly localization, we adopt the Area Under the Per-Region Overlap (AUPRO), which measures the quality of pixel-level anomaly segmentation under controlled false positive rates. Specifically, instead of using a relatively loose integration threshold, we report AUPRO under stricter false positive rate (FPR) thresholds of 0.1 and 0.05, denoted as AUPRO@10\% and AUPRO@5\%, respectively. These tighter thresholds better reflect practical industrial requirements, where excessive false alarms are undesirable. To provide a comprehensive evaluation, we present the performance for each individual class. However, since our primary objective is to assess the overall effectiveness of the proposed method in distributed industrial settings, we place greater emphasis on the mean performance across all classes.

\subsection{Online Data Generation}
In our experiments, the online learning setting requires the training dataset to evolve dynamically, with new data collected at the beginning of each global round. To this end, we adapt the original datasets to support distributed and online processing. Given the similarity in dataset characteristics and scale, we adopt a consistent online data generation strategy for both datasets. In the following, we take MVTec 3D-AD as a representative example to illustrate the process.

For the MVTec 3D-AD dataset, we adopt the defect-free training split as the long-term data source, while keeping the official validation and test splits unchanged for evaluation. The 10 object categories are distributed across 5 clients in a structured Non-IID manner, where each client is assigned 4 classes and each class is shared by exactly two clients. This overlapping assignment ensures collaborative learning while preserving distribution heterogeneity. Specifically, for each class, its normal samples are evenly partitioned between the two corresponding clients to construct client-specific long-term data sources. To simulate realistic online data arrival, we design a progressive streaming mechanism based on fixed-size packets. At each global round, every client receives a packet of at most 10 samples drawn from its remaining long-term data pool. The class composition of each packet is dynamically determined by a Dirichlet distribution, which introduces stochastic variability in class proportions across rounds and naturally reflects temporal distribution shifts. The sampling process also respects the remaining availability of each class to ensure that no class is oversampled. Unlike sliding-window settings, the local dataset evolves in an accumulative manner: newly arrived samples are continuously appended, and no historical data is discarded. As a result, the number of samples per class at each client gradually increases over time, forming a realistic online data growth process. The per-round class distribution is explicitly tracked to characterize the evolving local data statistics. This design jointly captures the distributed, Non-IID, and progressively evolving online characteristics of the MODIAD setting.

For the Eyecandies dataset, the 10 object categories are distributed across 5 clients in a structured Non-IID manner, where each client is assigned 4 classes and each class is shared by exactly two clients. Specifically, for each class, its normal samples are evenly partitioned between the two corresponding clients to construct client-specific long-term data sources. To simulate realistic online data arrival, we adopt a progressive streaming mechanism based on fixed-size packets. At each global round, every client receives a packet of at most 20 samples drawn from its remaining long-term data pool. The class composition of each packet is dynamically determined using a Dirichlet distribution, which introduces stochastic variability in class proportions across rounds and reflects temporal distribution shifts. The local dataset evolves in an accumulative manner, where newly arriving samples are continuously appended. Overall, this setup follows a design philosophy similar to that of the MVTec 3D-AD dataset.

\subsection{Model Detail}
In this section, we first describe the feature extraction components used to obtain modality representations, and then introduce the cross-modal mapping model designed to capture the intrinsic relationships among normal multimodal data.

\textbf{Feature Extractors}.
We use 2 Transformer-based feature extractors to separately extract the RGB feature and point clouds feature. For RGB features, we employ a Vision Transformer (ViT) \cite{han2022survey} to directly extract patch-level representations. To balance efficiency with detection granularity, we adopt the ViT-B/8 architecture. For improved performance, we utilize a ViT-B/8 model pretrained on ImageNet \cite{deng2009imagenet} using the DINO \cite{caron2021emerging} framework, which takes a 224 × 224 image as input and produces 784 patch features per image. As prior studies indicate that ViTs capture both global and local contextual information across layers, we use the 768-dimensional output from the final layer as the RGB feature representation for anomaly detection. For point cloud features, we adopt a Point Transformer \cite{zhao2021point} pretrained on the ShapeNet dataset \cite{chang2015shapenet} as our 3D feature extractor. Features from the 3rd, 7th, and 11th layers are used as the 3D representations. The Point Transformer first partitions the point cloud into point groups, analogous to image patches in a ViT, where each group is characterized by a central point representing its position and a set of neighboring points defining the group size.

\textbf{Cross-modal Mapping Models}. For each object category, we maintain an independent pair of cross-modal mapping networks to capture class-specific correspondences between RGB and point cloud modalities. Specifically, for each class $c$, we construct two mapping functions, $M_{c}^{2D \rightarrow 3D}$ and $M_{c}^{3D \rightarrow 2D}$, which transform 2D features into the 3D feature space and vice versa. This class-wise design allows the model to better adapt to heterogeneous appearance and geometric patterns across different industrial objects. Each mapping network is implemented as a lightweight multi-layer perceptron (MLP). Given an input feature, it is first projected to an intermediate hidden space with dimension $((d^{in} + d^{out})/2)$, followed by one or multiple non-linear transformation layers with GELU activation, and finally mapped to the target feature dimension. To enhance modeling capacity when needed, a deeper variant is adopted by stacking multiple projection layers, enabling more expressive non-linear cross-modal alignment. This design provides an efficient mechanism for learning bidirectional, class-specific feature correspondences while maintaining low computational overhead in distributed settings.

\subsection{Benchmarks}
In our experiments, we employ several baseline methods for performance comparison. As multimodal industrial anomaly detection in distributed and online settings has not been previously studied, standardized benchmarks are unavailable. Therefore, we construct a set of representative baselines to systematically evaluate the effectiveness of the proposed method. A detailed explanation of each benchmark is provided below.

\textbf{Random Scheduling (RS)} In this setting, each client randomly selects $\Gamma^{cop}_k $ classes from its locally available set and reports the corresponding client–class pairs to the server. The server then randomly selects a subset of size $M^{com}$ from the reported client–class pairs to participate in subsequent training.

\textbf{Sufficient Only (SO)} In this setting, each client selects  $\Gamma^{cop}_k$ classes to report based on its local data volume. The server then selects a subset of size $\Gamma^{com}$ from the reported client–class pairs, prioritizing those with larger data volumes, to participate in subsequent training.

\textbf{Balanced Only (BO)} In this setting, each client selects the $\Gamma^{cop}_k$ classes with the smallest $B_c^t$ values observed so far and reports them to the server. The server then selects $\Gamma^{com}$ client–class pairs with the smallest $B_c^t$ values from the reported set, thereby prioritizing under-updated classes and promoting balanced training across classes.

Based on the simulation setup described above, we report the experimental results in the following sections.

\subsection{Simulation Results}
In this section, we compare the performance of the proposed algorithm with baseline methods under the MODIAD setting.

\begin{table*}[t]
\centering
\caption{I-AUROC, AUPRO@10\% and AUPRO@5\% on MVTec 3D-AD for Performance Comparison (Round@50). Best results in \textbf{bold}, runner-ups \underline{underlined}.}
\label{tab:pc-mvtec}
\resizebox{\textwidth}{!}{
\begin{tabular}{c|c|cccccccccc|c}
\hline
\multirow{2}{*}{Metric} & \multirow{2}{*}{Method} 
& \multicolumn{10}{c|}{Class} & \multirow{2}{*}{Mean} \\
\cline{3-12}
 &  & Bagel & Cable & Carrot & Cookie & Dowel & Foam & Peach & Potato & Rope & Tire &  \\
\hline
\multirow{4}{*}{I-AUROC}
& RS   & 0.981 & 0.699 & \underline{0.972} & \textbf{0.992} & 0.919 & 0.811 & 0.913 & 0.855 & \underline{0.956} & 0.790 & 0.888 \\
& SO    & 0.703 & \textbf{0.850} & 0.550 & 0.988 & \underline{0.940} & \underline{0.855} & \underline{0.943} & \underline{0.882} & 0.951 & \textbf{0.871} & 0.853 \\
& BO   & \textbf{0.985} & 0.805 & 0.971 & \underline{0.991} & 0.908 & 0.801 & 0.939 & 0.865 & 0.955 & 0.776 & \underline{0.899} \\
& SMG   & \underline{0.983} & \underline{0.819} & \textbf{0.976} & 0.990 & \textbf{0.956} & \textbf{0.864} & \textbf{0.952} & \textbf{0.897} & \textbf{0.967} & \underline{0.869} & \textbf{0.927} \\
\hline

\multirow{4}{*}{AUPRO@10\%}
& RS   & 0.933 & 0.638 & \underline{0.945} & \underline{0.892} & 0.783 & \underline{0.858} & 0.929 & \underline{0.944} & \underline{0.917} & 0.912 & 0.875 \\
& SO    & 0.795 & \textbf{0.866} & 0.795 & 0.886 & \underline{0.798} & \textbf{0.881} & \underline{0.934} & \underline{0.944} & 0.913 & \underline{0.936} & 0.874 \\
& BO   & \underline{0.936} & 0.808 & \underline{0.945} & \textbf{0.894} & 0.777 & 0.853 & \underline{0.934} & \textbf{0.945} & \underline{0.917} & 0.894 & \underline{0.890} \\
& SMG   & \textbf{0.937} & \underline{0.849} & \textbf{0.946} & \textbf{0.894} & \textbf{0.807} & \textbf{0.881} & \textbf{0.935} & \textbf{0.945} & \textbf{0.920} & \textbf{0.929} & \textbf{0.904} \\
\hline

\multirow{4}{*}{AUPRO@5\%}
& RS   & \underline{0.870} & 0.516 & \underline{0.891} & \underline{0.836} & 0.668 & 0.766 & 0.860 & \underline{0.889} & \underline{0.851} & 0.835 & 0.798 \\
& SO   & 0.621 & \textbf{0.764} & 0.596 & 0.828 & \underline{0.686} & \underline{0.788} & \underline{0.870} & 0.888 & 0.845 & \textbf{0.873} & 0.776 \\
& BO   & \textbf{0.876} & 0.687 & 0.890 & \textbf{0.838} & 0.661 & 0.755 & 0.869 & \underline{0.889} & 0.850 & 0.806 & \underline{0.812} \\
& SMG  & \textbf{0.876} & \underline{0.740} & \textbf{0.893} & \textbf{0.838} & \textbf{0.698} & \textbf{0.791} & \textbf{0.871} & \textbf{0.890} & \textbf{0.856} & \underline{0.862} & \textbf{0.832} \\
\hline

\end{tabular}
}
\end{table*}

\begin{table*}[t]
\centering
\caption{I-AUROC, AUPRO@10\% and AUPRO@5\% on Eyecandies for Performance Comparison (Round@50). Best results in \textbf{bold}, runner-ups \underline{underlined}.}
\label{tab:pc-eye}
\resizebox{\textwidth}{!}{
\begin{tabular}{c|c|cccccccccc|c}
\hline
\multirow{2}{*}{Metric} & \multirow{2}{*}{Method} 
& \multicolumn{10}{c|}{Class} & \multirow{2}{*}{Mean} \\
\cline{3-12}
 &  & Cane & Choco & Pralin & Confet & Gummy & Hazel & Licor & Lolli & Marsh & Pepper &  \\
\hline
\multirow{4}{*}{I-AUROC}
& RS   & 0.500 & 0.741  & 0.800  & \textbf{0.856}  & 0.809  & \textbf{0.726}  & 0.810  & 0.814  & \textbf{0.971}  & \underline{0.838}  & 0.786  \\
& SO    & \textbf{0.584} & \underline{0.892}  & 0.653  & 0.803  & \textbf{0.850}  & 0.702  & \underline{0.811}  & \textbf{0.829}  & 0.958  & 0.824  & 0.790  \\
& BO   & 0.478 & 0.886  & \underline{0.801}  & 0.828  & \underline{0.826}  & 0.719  & 0.807 & \underline{0.826} & 0.955  & \textbf{0.840}  & \underline{0.796} \\
& SMG & \underline{0.503} & \textbf{0.896} & \textbf{0.818}  & \underline{0.850}  & 0.798  & \underline{0.723}  & \textbf{0.845} & 0.822  & \underline{0.960} & 0.812 & \textbf{0.803}   \\
\hline

\multirow{4}{*}{AUPRO@10\%}
& RS  & \textbf{0.826} & 0.682 & \textbf{0.652} & 0.852  & \textbf{0.718}  & 0.528  & \underline{0.650} & 0.705  & 0.895  & \underline{0.824}  & 0.733  \\
& SO   & 0.766 & \textbf{0.813}  & 0.175  & \textbf{0.870}  & 0.707  & \underline{0.543}  & 0.644  & \underline{0.717}  & \underline{0.896}  & \textbf{0.825}  & 0.695  \\
& BO   & 0.806 & 0.807  & 0.629  & 0.848  & \underline{0.717}  & 0.509  & 0.639  & 0.710 & 0.892  & 0.817  & \underline{0.737}  \\
& SMG   & \underline{0.813} & \underline{0.809}  & \underline{0.645}  & \underline{0.864}  & 0.713  & \textbf{0.561} & \textbf{0.659}  & \textbf{0.721}  & \textbf{0.897}  & 0.819  & \textbf{0.750} \\
\hline

\multirow{4}{*}{AUPRO@5\%}
& RS   & \textbf{0.655} & 0.527  & \textbf{0.573} & 0.753  & \underline{0.628}  & 0.416  & 0.577  & 0.502  & \underline{0.840} & \underline{0.736} & 0.620  \\
& SO   & 0.539 & \textbf{0.748} & 0.092  & \textbf{0.767}  & 0.621  & \underline{0.434}  & 0.579  & \underline{0.519}  & 0.838  & \textbf{0.739}  & 0.587 \\
& BO   & 0.615 & 0.737  & 0.555  & 0.739  & \textbf{0.631} & 0.400 & \underline{0.581}  & 0.517  & 0.835  & 0.727  & \underline{0.633}  \\
& SMG   & \underline{0.629} & \underline{0.743}  & \underline{0.566}  & \underline{0.763} & 0.626  & \textbf{0.436}  & \textbf{0.596}  & \textbf{0.521}  & \textbf{0.841}  & 0.732  & \textbf{0.645} \\
\hline

\end{tabular}
}
\end{table*}

\begin{figure}[htp]
\vspace{-5pt}
\centering
\subfloat{\includegraphics[width=0.9\linewidth]{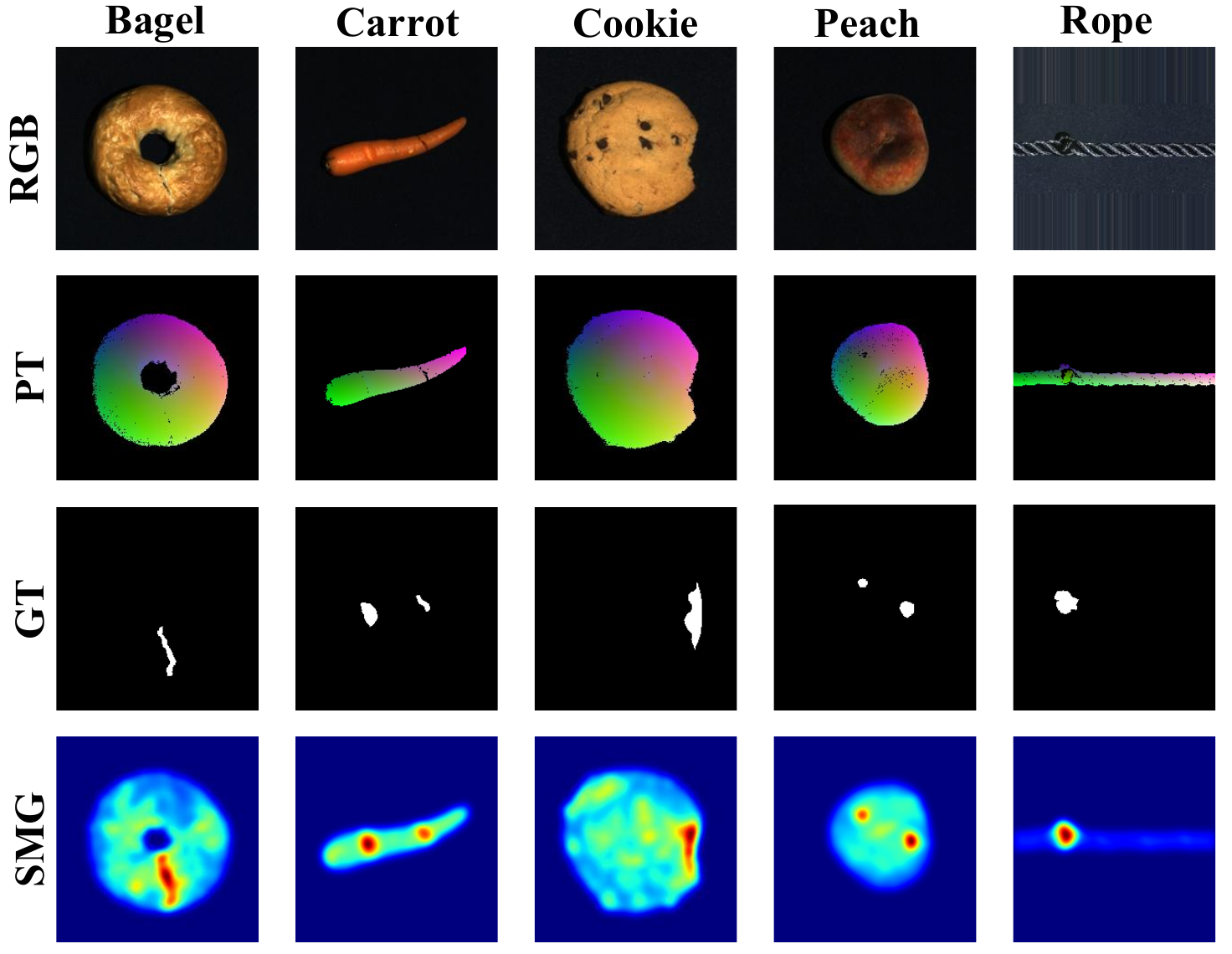}} 
\caption{Visualization Localization Results (MVTec 3D-AD)}
\label{vis-re}
\vspace{-5pt}
\end{figure}

\textbf{Performance Comparison}. We begin by evaluating the learning performance of the proposed SMG algorithm compared to benchmarks in the MODIAD scenario. We evaluate the proposed method over 50 consecutive rounds of cross-class multimodal normal data. The final anomaly detection performance, including I-AUROC, AUPRO@10\% , and AUPRO@5\% , is reported at both the class level and the overall mean in Table~\ref{tab:pc-mvtec} and Table~\ref{tab:pc-eye} for the MVTec 3D-AD and Eyecandies datasets, respectively. Based on the above figures, we get several key observations. SMG achieves significantly higher mean scores across all three metrics compared to the baseline methods RS, SO, and BO. Although SMG may slightly underperform on certain individual classes, the performance gaps are generally marginal. In contrast, SO, which selects client–class pairs solely based on data volume, tends to produce highly uneven results, achieving strong performance for some classes while performing poorly for others. BO, on the other hand, focuses primarily on average performance and disregards data availability, often selecting client–class pairs with limited data and consequently yielding inferior results compared to SMG. Meanwhile, RS exhibits only moderate performance due to its random selection strategy and further underperforms compared to BO. These observations demonstrate the effectiveness of the SMG algorithm and highlight the importance of balancing data sufficiency and class balance in multi-class anomaly detection scenarios. Notably, consistent trends are observed across both datasets. In Fig.~\ref{vis-re}, we present representative visualization results on the MVTec 3D-AD dataset, which facilitate qualitative evaluation of anomaly localization and clearly demonstrate the effectiveness of the proposed method.

\begin{figure}[htbp]
\centering
\vspace{-10pt}

\subfloat[MVTec 3D-AD Dataset]{
\includegraphics[width=0.9\linewidth]{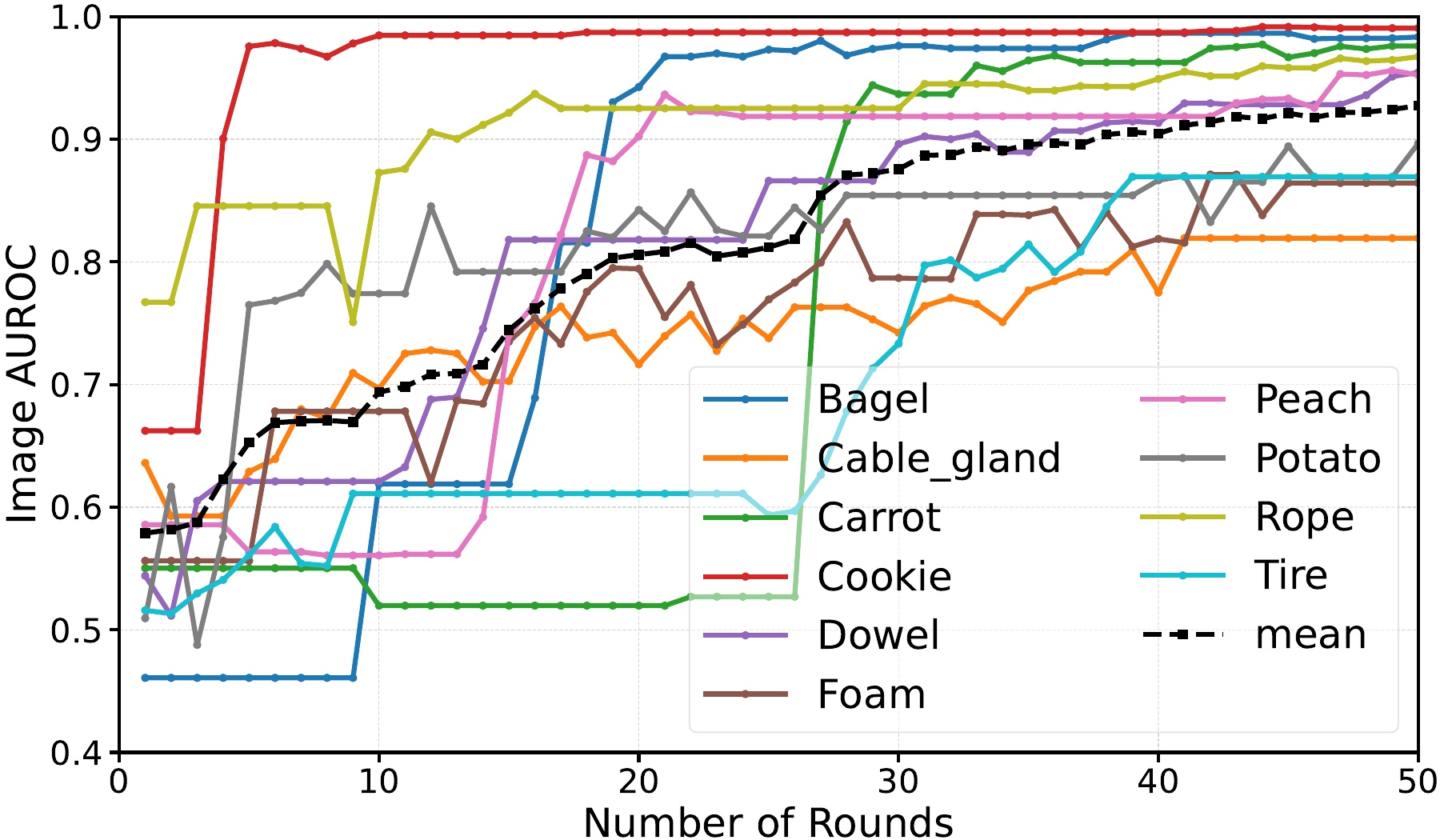}
} \\
\subfloat[Eyecandies Dataset]{
\includegraphics[width=0.9\linewidth]{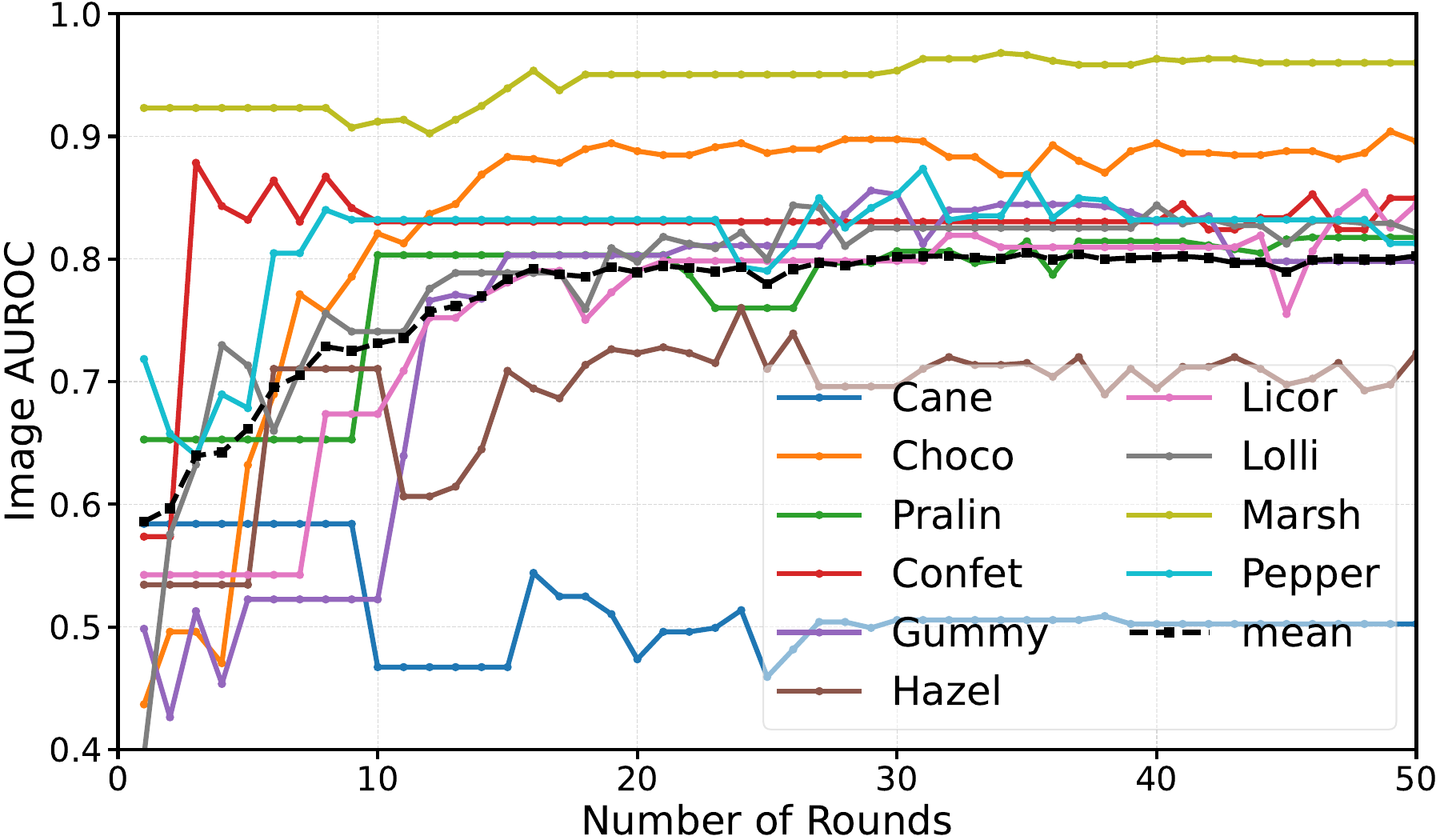}
}
\caption{Class-Wise Temporal Performance Evolution (SMG) }
\label{temp-pe}
\vspace{-10pt}
\end{figure}

\textbf{Class-Wise Temporal Performance Evolution}. In this section, we present a detailed analysis of the temporal evolution of detection performance, including both class-wise results and the overall mean. The results for the MVTec 3D-AD and Eyecandies datasets are illustrated in Fig.~\ref{temp-pe}(a) and Fig.~\ref{temp-pe}(b), respectively. We first observe that the I-AUROC trends vary noticeably across different classes. Taking the MVTec 3D-AD dataset as an example, some classes, such as “cookie,” rapidly achieve high detection performance, whereas others, such as “cable,” exhibit more gradual improvement before reaching satisfactory performance. Despite these class-wise variations, the overall detection performance consistently improves as online data accumulates and training progresses. In particular, the mean detection performance shows a steady upward trend across rounds, demonstrating the effectiveness of the SMG based distributed industrial anomaly detection framework in improving overall detection performance.

\begin{figure}[h]
\centering
\vspace{-10pt}

\subfloat[Cumulative Communication Cost (MVTec 3D-AD)]{
\includegraphics[width=0.9\linewidth]{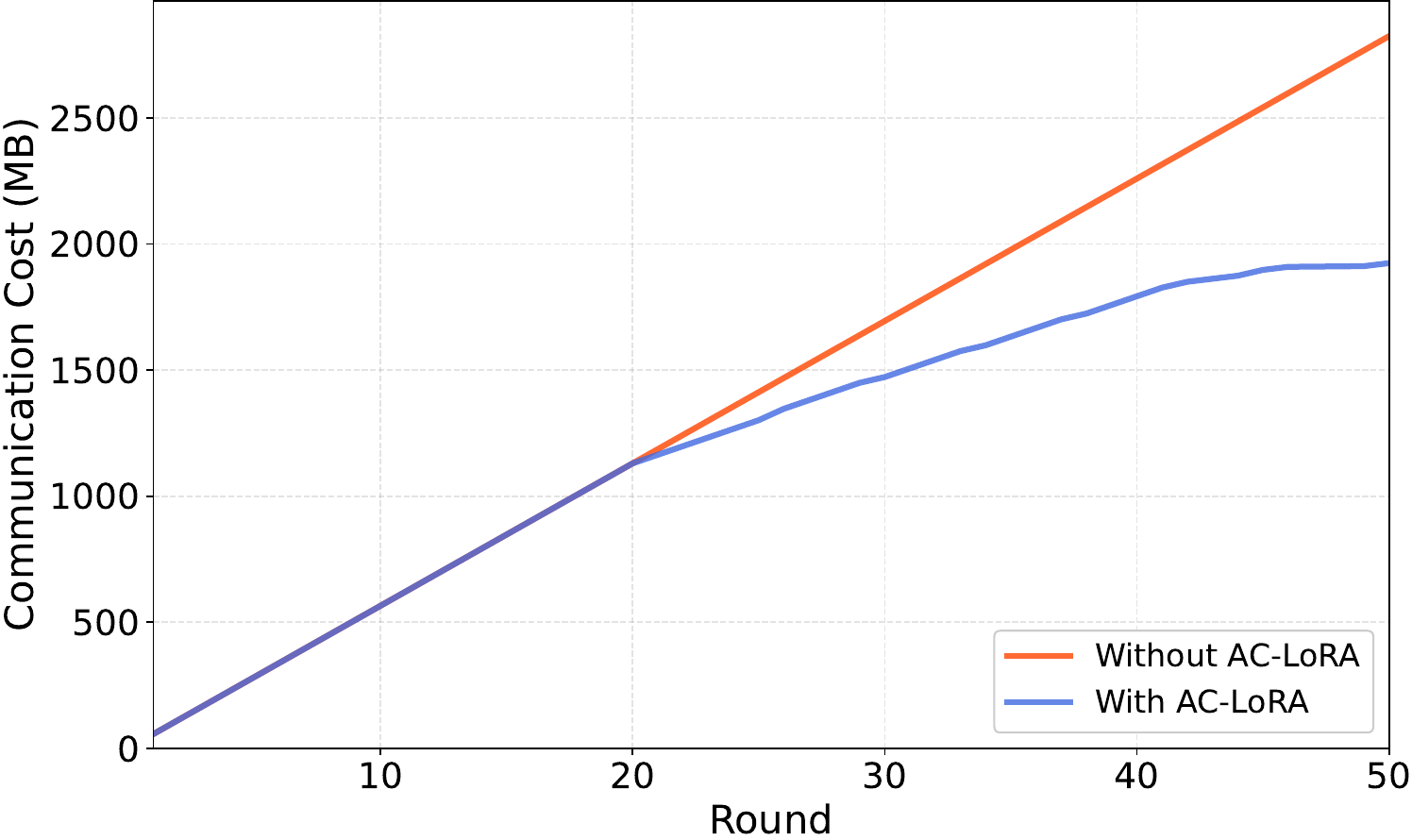}
} \\
\subfloat[Cumulative Communication Cost (Eyecandies)]{
\includegraphics[width=0.9\linewidth]{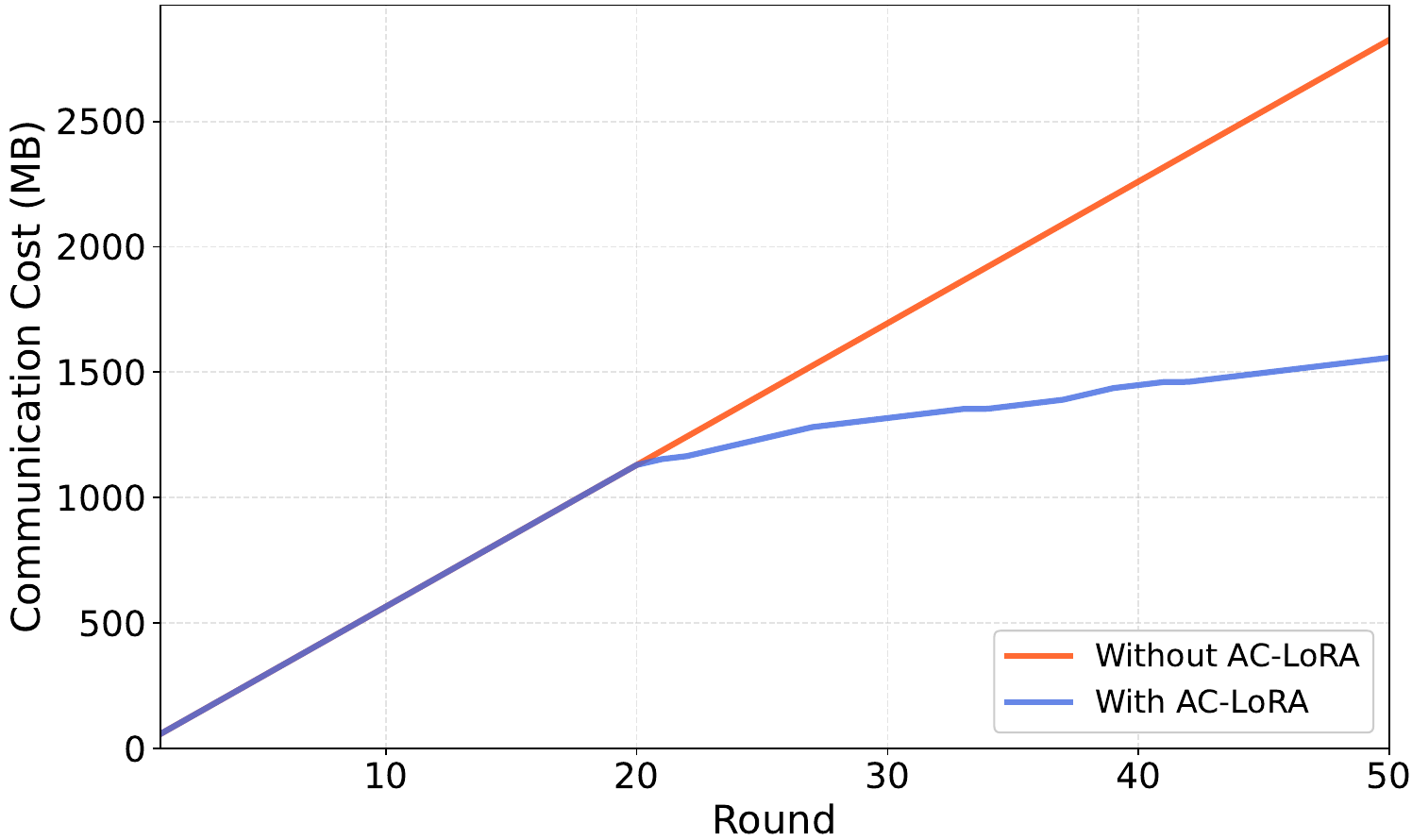}
}
\caption{Communication Cost with and without REC-LoRA . (Computational cost follows a similar trend.)}
\label{lora-cc}
\vspace{-10pt}
\end{figure}

\textbf{Performance Evolution of REC-LoRA}. We evaluate the learning performance of MODIAD under the proposed resource efficient  class-wise low-rank adaptation strategy. The final anomaly detection performance, including I-AUROC, AUPRO@10\% , and AUPRO@5\% , is reported at both the class level and the overall mean in Table~\ref{tab:lora-mvtec} and Table~\ref{tab:lora-eye} for the MVTec 3D-AD and Eyecandies datasets, respectively. In addition, Fig.~\ref{lora-cc}(a) and Fig.~\ref{lora-cc}(b) compare the communication and computational costs with and without REC-LoRA on the both datasets. From the above results, we observe that REC-LoRA achieves a favorable trade-off between detection performance and computational and communication overhead. Overall, REC-LoRA effectively reduces system overhead with only minimal impact on detection accuracy, thereby addressing key deployment challenges in edge-based distributed industrial anomaly detection.

\begin{table*}[t]
\centering
\caption{I-AUROC, AUPRO@10\% and AUPRO@5\% on MVTec 3D-AD for Performance Evolution of REC-LoRA (Round@50, Rank@32). Best results in \textbf{bold}.}
\label{tab:lora-mvtec}
\resizebox{\textwidth}{!}{
\begin{tabular}{c|c|cccccccccc|c}
\hline
\multirow{2}{*}{Metric} & \multirow{2}{*}{Case} 
& \multicolumn{10}{c|}{Class} & \multirow{2}{*}{Mean} \\
\cline{3-12}
 &  & Bagel & Cable & Carrot & Cookie & Dowel & Foam & Peach & Potato & Rope & Tire &  \\
\hline
\multirow{2}{*}{I-AUROC}
& w/o REC-LoRA   & \textbf{0.983} & \textbf{0.819} & \textbf{0.976} & \textbf{0.990} & \textbf{0.956} & \textbf{0.864} & \textbf{0.952} & \textbf{0.897} & \textbf{0.967} & 0.869 & \textbf{0.927}  \\
& w/ REC-LoRA   & 0.964 & 0.811  & 0.948  & 0.986 & 0.891 & 0.855  & 0.932  & 0.874  & 0.948  & \textbf{0.870}  & 0.908  \\
\hline

\multirow{2}{*}{AUPRO@10\%}
&  w/o REC-LoRA   & \textbf{0.937} & 0.849 & \textbf{0.946} & \textbf{0.894} & \textbf{0.807} & 0.881& \textbf{0.935} & \textbf{0.945} & \textbf{0.920} & \textbf{0.929} & \textbf{0.904}  \\
&  w/ REC-LoRA     & 0.930 & \textbf{0.851}  & 0.943  & 0.886  & 0.749  & \textbf{0.885}  & 0.928  & 0.944 & 0.913  & 0.926  & 0.896  \\
\hline

\multirow{2}{*}{AUPRO@5\%}
&  w/o REC-LoRA   &  \textbf{0.876} & 0.740 & \textbf{0.893} & \textbf{0.838} & \textbf{0.698} & 0.791 & \textbf{0.871} & \textbf{0.890} & \textbf{0.856} & \textbf{0.862} & \textbf{0.832}  \\
&  w/ REC-LoRA     & 0.874 & \textbf{0.744} & 0.886  & 0.825 & 0.629  & \textbf{0.793}  & 0.850  & 0.888  & 0.843  & 0.860  & 0.819  \\
\hline

\end{tabular}
}
\end{table*}

\begin{table*}[t]
\centering
\caption{I-AUROC, AUPRO@10\% and AUPRO@5\% on Eyecandies for Performance Evolution of REC-LoRA (Round@50, Rank@32). Best results in \textbf{bold}.}
\label{tab:lora-eye}
\resizebox{\textwidth}{!}{
\begin{tabular}{c|c|cccccccccc|c}
\hline
\multirow{2}{*}{Metric} & \multirow{2}{*}{Case} 
& \multicolumn{10}{c|}{Class} & \multirow{2}{*}{Mean} \\
\cline{3-12}
 &  & Cane & Choco & Pralin & Confet & Gummy & Hazel & Licor & Lolli & Marsh & Pepper &  \\
\hline
\multirow{2}{*}{I-AUROC}
& w/o REC-LoRA  & \textbf{0.503} & \textbf{0.896} & \textbf{0.818}  & \textbf{0.850}  & 0.798  & \textbf{0.723}  & \textbf{0.845} & 0.822  & 0.960 & 0.812 & \textbf{0.803}  \\
& w/ REC-LoRA   & 0.494 & 0.894  & 0.765  & 0.835  & \textbf{0.845}  & 0.717  & 0.811  & \textbf{0.827}  & \textbf{0.963}  & \textbf{0.819}  & 0.797  \\
\hline

\multirow{2}{*}{AUPRO@10\%}
&  w/o REC-LoRA  & \textbf{0.813} & \textbf{0.809}  & \textbf{0.645}  & 0.864  & \textbf{0.713}  & 0.561 & \textbf{0.659}  & \textbf{0.721}  & \textbf{0.897}  & \textbf{0.819}  & \textbf{0.750}  \\
&  w/ REC-LoRA     & 0.799 & 0.796  & 0.619  & \textbf{0.866}  & \textbf{0.713}  & \textbf{0.562}  & 0.654  & 0.704  & 0.893 & 0.771 & 0.738 \\
\hline

\multirow{2}{*}{AUPRO@5\%}
&  w/o REC-LoRA    & \textbf{0.629} & \textbf{0.743}  & \textbf{0.566}  & 0.763 & \textbf{0.626}  & \textbf{0.436}  & \textbf{0.596}  & \textbf{0.521}  & \textbf{0.841}  & \textbf{0.732}  & \textbf{0.645}  \\
&  w/ REC-LoRA     & 0.604 & 0.721 & 0.541 & \textbf{0.776} & 0.623  & 0.430  & 0.580  & 0.508  & 0.835  & 0.682 & 0.630  \\
\hline

\end{tabular}
}
\end{table*}

Based on the above performance comparisons, which validate the effectiveness of the proposed algorithms under the MODIAD setting, we further conduct a series of ablation studies to examine the impact of key parameters on learning performance. All subsequent ablation results are obtained on the MVTec 3D-AD dataset.

\textbf{Impact of Computational Capacity $ \Gamma^{cop}_k$}. In this section, we investigate the impact of computational capacity $\Gamma^{cop}_k$ on the learning performance of the SMG algorithm under the MODIAD setting with parameter $(\Gamma^{com} =5, \alpha = 0.5, \beta=0.5)$. The results on the MVTec 3D-AD dataset are reported in Table~\ref{tab:computation}, where we present the mean performance values. We observe that as the local computational capacity $ \Gamma^{cop}_k$ increases, more class-specific models can be selected for updating and aggregation in each round, leading to improved anomaly detection performance. Specifically, when $ \Gamma^{cop}_k$ increased from 1 to 2, all three detection metrics exhibit significant improvements. However, further increasing $\Gamma^{cop}_k$ from 2 to 3 yields limited performance gains. Although each client can handle more classes locally, the overall improvement is constrained by communication bottlenecks, as the global communication capacity $\Gamma^{com} =5$ remains fixed.

\begin{table}[t]
\centering
\caption{Ablation Study on Computational Capacity.}
\label{tab:computation}
\renewcommand{\arraystretch}{1.2}
\setlength{\tabcolsep}{8pt}
\begin{tabular}{c||c c c }
\hline
\textbf{$ M^{cop}_k$} & \textbf{I-AUROC} & \textbf{AUPRO@10\%} & \textbf{AUPRO@5\%} \\
\hline
1 & 0.912 & 0.898 & 0.823  \\
2 & \textbf{0.927} & 0.904 & \textbf{0.832}  \\
3 & \textbf{0.927} & \textbf{0.905} & \textbf{0.832}  \\
\hline
\end{tabular}
\end{table}

\textbf{Impact of Communicational Capacity $\Gamma^{com}$}. In this section, we investigate the impact of communicational capacity $\Gamma^{com}$ on the learning performance of the SMG algorithm under the MODIAD setting with parameter $(\Gamma^{cop}_k =2, \alpha = 0.5, \beta=0.5)$. The results on the MVTec 3D-AD dataset are reported in Table~\ref{tab:communication}, where we present the mean performance values. It is observed that increasing the global communication capacity $\Gamma^{com}$ from 3 to 5 yields a significant performance improvement across all three evaluation metrics. However, further increasing $\Gamma^{com}$ from 5 to 7 results in only marginal gains. This is because, with $\Gamma^{cop}_k =2$, there are at most 10 candidate client–class pairs available in each round. The SMG algorithm selects pairs based on maximum marginal gain, and once the most informative pairs have been chosen, additional selections contribute limited benefit. Moreover, without a corresponding increase in the clients’ computational capacity or the availability of new candidate pairs, simply enlarging $\Gamma^{com}$ does not substantially improve overall performance.

\begin{table}[t]
\centering
\caption{Ablation Study on Communicational Capacity.}
\label{tab:communication}
\renewcommand{\arraystretch}{1.2}
\setlength{\tabcolsep}{8pt}
\begin{tabular}{c||c c c }
\hline
\textbf{$M^{com}$} & \textbf{I-AUROC} & \textbf{AUPRO@10\%} & \textbf{AUPRO@5\%} \\
\hline
3 & 0.908 & 0.893 & 0.816   \\
5 & 0.927  & 0.904 & 0.832   \\
7 &  \textbf{0.928} &  \textbf{0.905} & \textbf{0.833}   \\
\hline
\end{tabular}
\end{table}

\textbf{Impact of REC-LoRA rank $r$}. In this section, we investigate the impact of rank $r$ on the detection performance of the proposed REC-LoRA strategy under the MODIAD setting. The results on the MVTec 3D-AD dataset are reported in Table~\ref{tab:rank}, where we present the mean performance values. There is a clear correlation between rank and detection performance. Specifically, for I-AUROC, performance consistently improves as the rank increases, whereas for AUPRO metrics, performance remains largely stable when the rank increases from 8 to 16, although variations across individual classes are still observed. These findings indicate that REC-LoRA is well-suited for MODIAD scenarios. In practice, the rank 
$r$ can be flexibly adjusted according to the communication and computational capabilities of clients, enabling an effective trade-off between detection performance and learning cost.

\begin{table}[t]
\centering
\caption{Ablation Study on Rank of REC-LoRA.}
\label{tab:rank}
\renewcommand{\arraystretch}{1.2}
\setlength{\tabcolsep}{8pt}
\begin{tabular}{c||c c c }
\hline
\textbf{$r$} & \textbf{I-AUROC} & \textbf{AUPRO@10\%} & \textbf{AUPRO@5\%} \\
\hline
8 & 0.875 &  0.890 &  0.810   \\
16 &  0.882  &  0.890 &  0.810  \\
32 &   \textbf{0.908} &   \textbf{0.896} &  \textbf{0.819}  \\
\hline
\end{tabular}
\end{table}

\section{Conclusion}
In this work, we introduce the concept of Multimodal Online Distributed Industrial Anomaly Detection (MODIAD) for the first time and provide a comprehensive introduction of its workflow and underlying principles. Compared with traditional FL frameworks based on supervised learning, distributed industrial anomaly detection is characterized by cross-class multi-model collaborative training and unsupervised learning objectives, making existing FL techniques unsuitable for direct application. Building upon the identified principles of data sufficiency and class balance, we formulate the Multi-class Intelligent Scheduling problem and propose a Sequential Marginal Gain Greedy solver to address it. Furthermore, considering the computational and communication constraints of edge devices, we design a Resource Efficient Class-Wise Low Rank Adaptation strategy to achieve parameter efficient learning. Extensive experiments conducted on two representative multimodal industrial anomaly detection datasets demonstrate that the proposed approach consistently outperforms baseline methods. For future work, from a practical systems perspective, we plan to establish a testbed environment for MODIAD to further investigate the challenges encountered in real-world deployment scenarios. From a theoretical perspective, we aim to develop a more advanced “one-for-all” framework based on a unified model, enabling unified representation learning, knowledge sharing, and adaptive scalability in MODIAD.

\bibliographystyle{IEEEtran}
\bibliography{bibligraphy}

\end{document}